\pdfoutput=1

\documentclass[11pt]{article}

\usepackage[]{acl}

\usepackage{amsthm,amsmath,amsfonts,bm,xspace}
\usepackage{color}

\def\eqref#1{(\ref{#1})}

\def\1{\bm{1}}

\def\va{{\bm{a}}}

\def\vx{{\bm{x}}}
\def\vy{{\bm{y}}}

\DeclareMathAlphabet{\mathsfit}{\encodingdefault}{\sfdefault}{m}{sl}
\SetMathAlphabet{\mathsfit}{bold}{\encodingdefault}{\sfdefault}{bx}{n}

\DeclareMathOperator*{\argmax}{arg\,max}

\usepackage{times}
\usepackage{latexsym}
\usepackage{graphicx}
\usepackage{array,booktabs}
\usepackage{enumitem}
\usepackage{arydshln}
\usepackage[T1]{fontenc}

\usepackage[utf8]{inputenc}

\usepackage{microtype}

\usepackage{inconsolata}

\definecolor{dgreen}{rgb}{0,0.55,0}
\definecolor{mgreen}{rgb}{0,0.7,0}

\title{Improving Cross-Domain Low-Resource Text Generation through LLM Post-Editing: A Programmer-Interpreter Approach}

\author{Zhuang Li, Levon Haroutunian, \\
    \textbf{
    Raj Tumuluri,
    Philip Cohen,
    Gholamreza Haffari }
    \\
         Openstream.ai\\ 
        \texttt{\{zhuang.li, levon, raj, phil.cohen, reza.haffari\}@openstream.com}
         }

\begin{document}
\abovedisplayskip=0.25pt
\abovedisplayshortskip=0.25pt
\belowdisplayskip=0.25pt
\belowdisplayshortskip=0.25pt
\maketitle
\begin{abstract}

Post-editing has proven effective in improving the quality of text generated by large language models (LLMs) such as GPT-3.5 or GPT-4, particularly when direct updating of their parameters to enhance text quality is infeasible or expensive. However, relying solely on smaller language models for post-editing can limit the LLMs' ability to generalize across domains. Moreover, the editing strategies in these methods are not optimally designed for text-generation tasks.  To address these limitations, we propose a neural programmer-interpreter approach that preserves the domain generalization ability of LLMs when editing their output. The editing actions in this framework are specifically devised for text generation. Extensive experiments demonstrate that the programmer-interpreter significantly enhances GPT-3.5's performance in logical form-to-text conversion and low-resource machine translation, surpassing other state-of-the-art (SOTA) LLM post-editing methods in cross-domain settings.

\end{abstract}

\section{Introduction}
\begin{figure}[ht]
    \centering
    \includegraphics[width=.7\columnwidth]{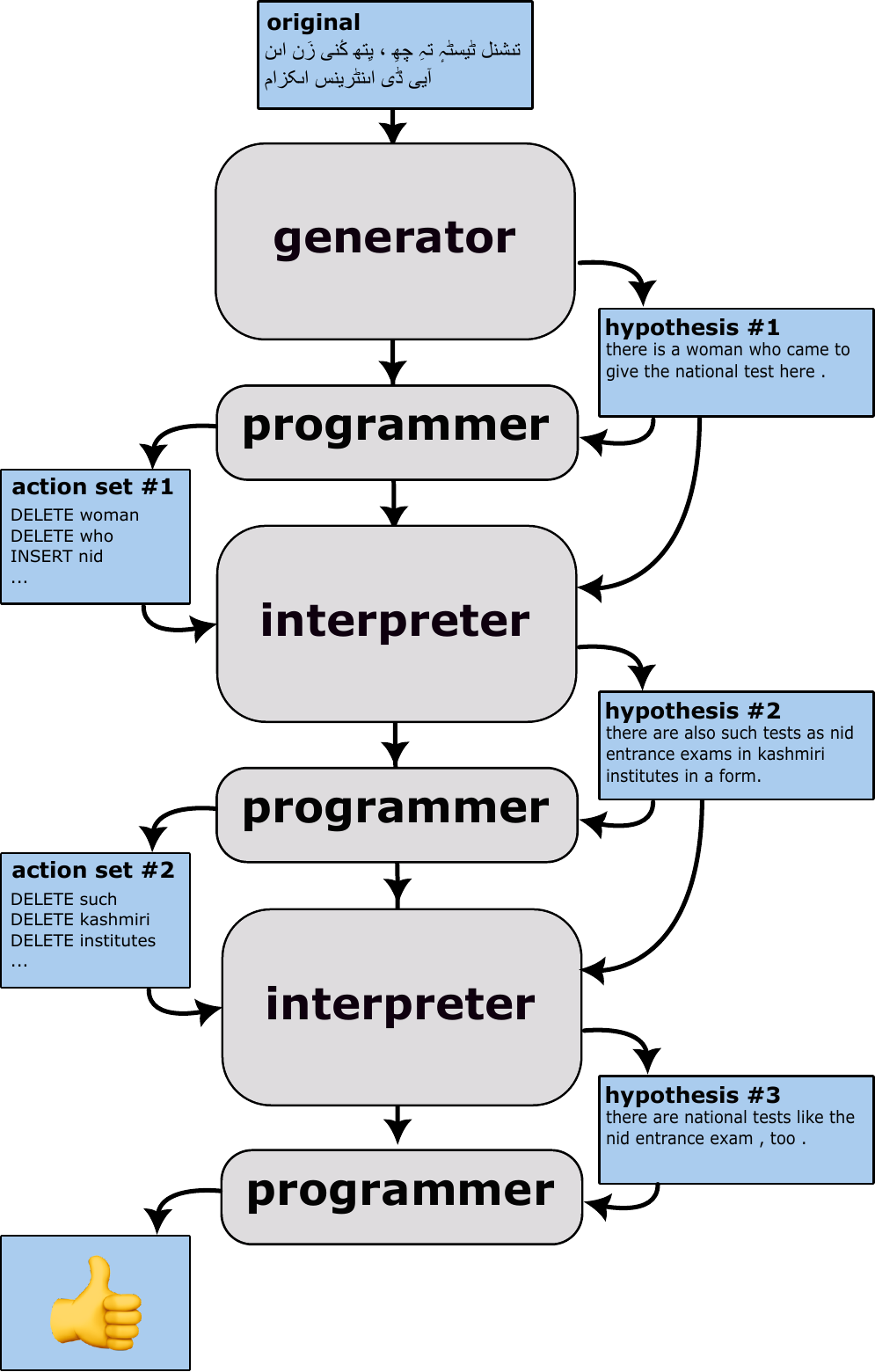}
    \vspace{-2mm}
    \caption{The diagram of our post-editing architecture.}
    \label{fig:diagram}
    \vspace{-2mm}
\end{figure}
Large pre-trained language models like GPT-3.5\footnote{https://platform.openai.com/docs/models/gpt-3-5-turbo} or GPT-4\footnote{https://platform.openai.com/docs/models/gpt-4-and-gpt-4-turbo} have gained significant attention in natural language research. However, fine-tuning these models for specific tasks is challenging due to limited computational resources or inaccessible parameters. Consequently, many researchers resort to using web APIs for instructing LLMs, leveraging zero-shot or few-shot in-context learning, enabling the LLMs to tackle tasks they weren't explicitly trained for. Unfortunately, this approach falls short when tackling some low-resource sequence generation tasks in machine translation (MT), and logical form (LF)-to-text translation, as shown in~\citet{lai2023chatgpt,haroutunian2023reranking}. In such cases, minimal task-specific data was available during the LLMs' pre-training phase. The output quality of LLMs for such tasks is compromised due to the absence of task-specific knowledge.

To address this challenge, a promising set of solutions suggests integrating task-specific knowledge into language models through post-editing the generated text using a smaller model fine-tuned on task-specific data. Yet, these methods are not without their drawbacks. Our findings indicate that exclusive reliance on a smaller model for editing, e.g. Self-Correct~\cite{welleck2022generating},  results in suboptimal performance in domain generalization scenarios, likely due to the inherently limited domain knowledge within these smaller models.


As LLMs (i.e. GPT-3.5 or GPT-4) have shown superior domain generalization ability~\cite{wang2023robustness,yang2023harnessing} over the fine-tuned model, we introduce an innovative approach based on the programmer-interpreter framework \cite{DBLP:journals/corr/ReedF15}, which benefits from the domain generalization ability from LLMs. The programmer component - a smaller language model fine-tuned on task-specific data - delivers precise edit instructions to the larger language model, thus infusing the large model with task-specific knowledge. The interpreter, in turn, edits the large model's output given the provided instructions. Contrary to the Self-Correct~\cite{welleck2022generating} approach that utilizes smaller, fine-tuned models for editing, our interpreter is also an LLM. The editing is accomplished through the use of prompts that include editing instructions, eliminating the need for any additional fine-tuning. This distinct framework guarantees the preservation of the LLM's domain generalization ability while simultaneously benefiting from the task-specific knowledge encoded by the programmer. Our method distinguishes itself from approaches like PiVe~\cite{han2023pive}, which also employ an LLM as the interpreter but focus on graph generation tasks. In contrast, our approach specifically designs word-level editing actions in the instructions, tailored to enhance text generation. This targeted strategy renders our method more effective for text-generation tasks.



Overall, our key contributions are as follows:

\begin{itemize}
    \item We introduce a novel programmer-interpreter method that enhances LLM in low-resource cross-domain text generation tasks. This approach capitalizes on the programmer's ability to encode task-specific knowledge and the interpreter's prowess in domain generalization.
    
    \item We design editing operations optimized for text generation tasks, leading to substantial text quality improvements by simply prompting the LLMs with action instructions.
    \item In scenarios where training and test data span different domains, our comprehensive empirical studies confirm that the method outperforms all existing LLM post-editing baselines in low-resource MT and LF-to-Text. 
\end{itemize}


\section{Programmer-Interpreter Approach}
The objective in LF-to-text and MT tasks using LLMs is to generate a high-quality output text \( \vy \), denoted as \( \vy' = \argmax_{\vy \in \mathcal{Y}} P(\vy|\vx, \mathcal{C}) \), given an input \( \vx \) (e.g., LF, source-language utterance) and an exemplar pool \( \mathcal{C} =\{(\vx_{j},\vy_{j},\vy^{*}_{j},\va^{*}_{j})\}^{|\mathcal{C}|}_{j=1} \). Here, \(\vx_i\) and \(\vy_j\) are the ground truth input-output pairs, \(\vy^{*}_j\) is the imperfect translation of \(\vx_i\), and \(\va^{*}_{j}\) represents the Oracle edit actions that can modify \(\vy^{*}_j\) into \(\vy_j\). Our approach focuses on achieving high-quality generation through iterative refinement of the initial output text produced by an LLM. Specifically, the iterative refinement framework includes three-parameterized modules: a Generator, a Programmer, and an Interpreter,\footnote{To save space, we simplify the marginalization notation.}
\begin{align}
&P(\vy^{t}  |\vx, \mathcal{C})  =   \overbrace{P(\vy^0|\vx,M(\cdot))}^{Generator} \times \\ 
    & \sum^{t-1}_{\{\va,\vy\}}  \prod^{t-1}_{i=0}  (\overbrace{P(\vy^{i+1}|\va^i,\vy^i,\vx, A(\cdot))}^{Interpreter} \times \overbrace{P(\va^i|\vy^i,\vx))}^{Programmer} 
\end{align}
\noindent The Generator corresponds to the LLM (e.g. GPT-3.5, GPT-4). It produces the initial output text, $\vy^{0}$, given the input $\vx$, a set of examples retrieved by the function $M(\vx, \mathcal{C})$ when performing in-context learning. The Programmer, a module that creates editing actions \(\va^{i}\) given \(\vx\) and the current imperfect output \(\vy^{i}\), is a pre-trained Sequence-to-Sequence~\cite{sutskever2014sequence} language model, such as mT5~\cite{xue2021mt5} or flan-T5~\cite{chung2022scaling}, fine-tuned on a synthetic dataset. The Interpreter, essentially also an LLM, refines the imperfect intermediate output $\vy^{i}$ by processing instructions that incorporate predicted editing actions and few-shot editing examples, retrieved via the function $A(\vx,\mathcal{C})$. Please note that the Programmer has much fewer parameters than the LLM used by the Generator and Interpreter. After several iterative refinements, we arrive at the final output $\vy^{t}$ generated by the LLM. During generation, we assume no access to the parameters of the LLMs but only obtain the output text by providing prompting instructions. The implementation details of each module are as follows:
\paragraph{Generator.} To generate the initial output, we supply a prompt composed of a few-shot set of exemplar pairs, denoted as $M(\vx, \mathcal{C})=\{(\vx_{j},\vy_{j})\}^{m}_{j=1}$, selected from a pool of reference pairs $\mathcal{C}$. This is accompanied by an instruction prompting the LLM to produce output $\vy^{0}$ based on the input $\vx$. The retrieval function identifies the closest pairs by calculating the cosine similarity of TF-IDF features between $\vx$ and other instances of $\vx$ in $\mathcal{C}$.
\paragraph{Programmer.} After obtaining the initial or intermediate output \(\vy^{i}\) from either the Generator or the Interpreter, we combine the input \(\vx\) and \(\vy^{i}\) into a single sequence and feed it to the Programmer to generate a sequence of edit actions \(\va^{i}\). We create a synthetic training set \(\mathcal{T}\), extracted from the example pool \(\mathcal{C}\), for fine-tuning the Programmer. Each pair in \(\mathcal{T}\) is defined as \((\vx_{concat}, \va^{*})\), where \(\vx_{concat}\) is the concatenated sequence of \(\vx\) and \(\vy^{*}\), serving as the input for the Programmer. The output \(\va^{*}\) is the sequence of Oracle edit actions, synthetically generated based on the reference pairs in \(\mathcal{C}\). For each reference \(\vy \in \mathcal{C}\), we calculate the word-level edit distance to the imperfect translation \(\vy^{*}\), generating intermediate edit actions. Only \textit{INSERT}-word and \textit{DELETE}-word actions are retained in the sequence, forming the final training sequence \(\va^{*}\) for the Programmer. If \(\vy^{*}\) is identical to the reference \(\vy\), the action is labeled as ``\textit{NoAction}'', indicating that no refinement is needed for that instance. Unlike PiVe, which generates the imperfect translation \(\vy^{*}\) by scrambling the original \(\vy\), we directly use the initial output \(\vy^{0}\) from the Generator as \(\vy^{*}\) in both \(\mathcal{C}\) and \(\mathcal{T}\). This approach enables the Programmer to learn an action distribution that more effectively corrects translation errors from LLMs. In practice, to address the rarity of ``\textit{NoAction}'' instances, we supplement these cases by creating augmented pairs, each consisting of two identical $\vy$ sequences.


\begin{table*}[t]
    \centering
             \resizebox{0.8\textwidth}{!}{%
    \begin{tabular}{l|cccccc|ccc}
    \toprule
   \multicolumn{1}{c|}{ } &\multicolumn{6}{c}{MT (Kashmiri to English)} & \multicolumn{3}{|c}{LF-to-Text (AMR to English)} \\
    \toprule
 \multicolumn{1}{c|}{ } & \multicolumn{3}{c}{GEN} & \multicolumn{3}{c}{CONV} & \multicolumn{3}{|c}{Bio-AMR} \\
    Method & BLEU & BERT & ChrF++ & BLEU & BERT & ChrF++ & BLEU & BERT & ChrF++ \\
    \midrule
      Fine-tuned mT5/flan-T5 & \textbf{16.58} & 89.32 & 41.77 & 13.19 & 88.83 & 33.03 & 9.27  & 87.90 & 41.06\\
            \midrule
      GPT-3.5  &  & & & & & \\
      \ \ \ \ Initial  &  9.21 & 87.29 & 34.30 & 5.92 & 87.24 & 26.23 &  9.63 & 88.57 &  43.98 \\
      \ \ \ \ Self-Correct   &13.11 &89.02 & 38.98 & 12.73 & 89.61 & 33.76  & 11.64 & \textbf{89.44} & 46.05\\
            \ \ \ \ Algo-Refine  & 8.40 & 86.92 & 39.66 & 6.29 & 87.31 & 32.21 & 7.72 & 86.64 & 43.39 \\
      \ \ \ \ Self-Refine   & 8.13 & 86.54 & 31.78 & 4.73 & 86.55 & 24.13  &  8.67 & 87.34  &  39.63  \\
\hdashline
      \ \ \ \ Prog-Refine (Zero-shot Act.)  & 13.81 & 88.58 & 39.00 & 12.09 & 89.41 & 33.41  & 11.43 & 89.30 & 45.44  \\
            \ \ \ \ Prog-Refine (Few-shot  Act.)  & 16.32& \textbf{90.36} & \textbf{42.44} & \textbf{14.78} & \textbf{90.19} & \textbf{35.48}  & \textbf{13.64} & 89.27 & \textbf{47.69}  \\
\hdashline
    \ \ \ \ Prog-Refine (ORACLE)  & 43.48 & 92.11 & 65.29 &42.42 & 93.00 & 42.42  & 27.77 & 90.01 & 52.86 \\

      \bottomrule
    \end{tabular}  
    }
\vspace*{-3mm}
    \caption{The main results of MT on GEN and CONV test sets, and LF-to-Text on Bio-AMR test set.}
    \vspace*{-3mm}
    \label{tab:main_result}
\end{table*}

\paragraph{Interpreter.} To edit the intermediate output \(\vy^{i}\), we engage the LLM in the Interpreter role by providing it with prompting instructions. Given the edit instructions \(\va^{i}\) and a pair \((\vy^{i}, \vx)\), the LLM can \textit{INSERT} or \textit{DELETE} words \textit{in order} to generate the modified text \(\vy^{i+1}\). We also incorporate a few-shot examples that demonstrate editing procedures, extracted from \(\mathcal{C}\) and denoted as \(A(\vx, \mathcal{C}) = \{(\vx_{j},\vy_{j},\vy^{*}_{j},\va^{*}_{j})\}_{j=1}^{n}\). These examples are selected based on the cosine similarity between the TF-IDF features of \(\vx\) and those in \(\mathcal{C}\). Furthermore, to mimic action prediction errors from the Programmer, we adopt an \textit{adversarial in-context learning} strategy, similar to the approach in \citet{zhuo2023robustness}. This involves corrupting the action sequence by deleting Oracle actions with a certain probability \(d\%\). If an action is not deleted, we swap it with other actions from \(\mathcal{C}\) at the same probability \(d\%\). Through this manipulation, we have discovered that the LLM's exceptional text generalization ability enables it to effectively comprehend the editing instructions. As a result, it can generate high-quality text after performing the necessary edits, even if the predicted actions from the Programmer are not completely accurate. See Figures~\ref{fig:example_zeroshot} and~\ref{fig:examplar_adv} in the Appendix for zero/few-shot instruction examples.

\section{Experiments}

\paragraph{Setup.} In our experiments, we default to using GPT-3.5-turbo-0301 as the LLM for the Generator in both the zero-shot and few-shot settings. For the Interpreter, we use GPT-3.5-turbo-0301 in the zero-shot setting and GPT-3.5-turbo-16k\footnote{https://platform.openai.com/docs/models/gpt-3-5-turbo} in the few-shot setting. For the Generator used across all settings and baselines, we consistently use 0 and 5 shots for MT and LF-to-Text, respectively. For the Interpreter in the few-shot setting, we apply 10 and 5 action examples for MT and LF-to-Text, respectively, with a 50\% action corruption probability. For the MT and LF-to-Text tasks, we employ mT5-base and flan-T5-base as the backbones of the Programmers, respectively. These backbone choices are driven by a computationally efficient setup, ensuring the models fit within an Nvidia V100 with 16GB memory. We train our programmers with a dev set to select the optimal model. Our search for the best learning rate includes [5e-5, 1e-4, 2e-4], while the range of epochs considered is [5, 10, 20], with batch sizes 4. GPTs require no fine-tuning. Each generation of 1096 tokens costs approximately \$0.0015. We limit Self-Correct and Self-Refine to five editing iterations, as their performance typically stabilizes within this range. Conversely, Prog-Refine and Algo-Refine may require additional iterations for convergence, especially when `NoAction' instances are infrequent. Thus, we continue for up to 15 iterations, ceasing only if over 95\% of actions are `NoAction'. 


\paragraph{Datasets.}
To simulate low-data scenarios, in the context of \textbf{MT}, we utilize a Kashmiri-English dataset from IndicTrans2~\cite{gala2023indictrans2}. Since Kashmiri is a notably low-resource language, translating it poses a formidable challenge for LLMs. The dataset provides 26,016 training pairs, which we use to generate synthetic data for action generation. The development set consists of 997 pairs. The dataset includes two distinct test sets, GEN and CONV, with 1,024 and 1,503 pairs, respectively. Each of the training, development, and test sets originates from different domains. For \textbf{LF-to-Text}, we employ the AMR-LDC2.0\footnote{https://catalog.ldc.upenn.edu/LDC2017T10} dataset, which contains 22,550 AMR-English pairs for training and 1,368 pairs for development. For testing, we turn to a separate dataset, Bio-AMR\footnote{https://amr.isi.edu/download.html}, which offers 500 pairs in a different domain. Likewise, the AMR-to-Text task poses a low-resource challenge for LLMs.






\paragraph{Baselines.} We evaluate our approach, Prog-Refine, which utilizes zero-shot action exemplars (Zero-shot Act.) and few-shot action exemplars (Few-shot Act.) for Interpreters, against five baseline methods and an ORACLE setting 

i) \textbf{Fine-tuned Models} include mT5-base for MT and flan-T5-base for LF-to-Text generation, both of which are fine-tuned on the training set consisting of pairs $(\vx,\vy) \in \mathcal{C}$. These baseline models do not perform any refinement.

ii) \textbf{GPT-3.5 + Initial} simply applies the GPT-3.5 as the Generator to obtain the text without any further refinement.

iii) \textbf{GPT-3.5 + Self-Correct~\cite{welleck2022generating}} fine-tunes smaller models to be the Interpreter, fixing the output errors of the large models given the feedback. Here, we supply the edit actions produced by our Programmer as feedback to the fine-tuned Interpreters. These Interpreters are also built upon mT5-base or flan-T5-base. 

iv) \textbf{GPT-3.5 + Algo-Refine} directly `Insert' or `Delete' specific words in certain positions of the generated text instead of using an Interpreter to rewrite. Therefore, in this baseline, we also apply the Interpreter to predict the indices of words for actions. This method is prevalent in the MT literature; e.g. see ~\citet{vu2018automatic}.

v) \textbf{GPT-3.5 + Self-Refine~\cite{madaan2023self}} leverages an LLM to provide feedback for its own output, enabling self-refinement without the need for additional fine-tuning.

vi) \textbf{GPT-3.5 + Prog-Refine (ORACLE)} applies the ORACLE actions generated by comparing the reference in the test set with the initial output of the Generator, allowing for optimal refinement after one iteration in the Zero-shot Act. setting.

\paragraph{Evaluation Metrics.}
For LF-to-Text and MT tasks, we utilize three evaluation metrics to assess the quality of the final output text generated by the Programmer-Interpreter framework: BLEU~\cite{papineni2002bleu}, BERTScore~\cite{zhangbertscore} and Chrf++~\cite{popovic2017chrf++}. 



\subsection{Main Results and Analysis}

Table~\ref{tab:main_result} shows that \textit{GPT-3.5 + Prog-Refine} notably boosts the Generator's performance (i.e., \textit{GPT-3.5 + Initial}), underlining our method's effectiveness in cross-domain scenarios by enhancing initial GPT-3.5 outputs. Moreover, the few-shot setting (Few-shot Act.) significantly outperforms both the zero-shot (Zero-shot Act.) setting and all other refinement baselines. It's also noteworthy that applying ORACLE action to our method can lead to a roughly 30-point increase in BLEU score, suggesting substantial potential for improvement in our approach. In comparison, \textit{Self-Refine} shows minimal improvement, possibly due to its limited integration of task-specific knowledge. \textit{Algo-Refine} inconsistently improves the initial text, lacking the robustness seen in our method. We note that rewriting Interpreters, as in our approach and Self-Correct, can eliminate invalid actions, thus enhancing editing quality. However, \textit{Algo-Refine} does not possess this capability and is susceptible to incorrect feedback actions. The \textit{Self-Correct} method, using a fine-tuned Interpreter, along with fine-tuned mT5/flan-T5 models, demonstrates better performance than other baselines across various tasks. This underscores the importance of learning task-specific knowledge, especially in low-resource scenarios. Nonetheless, these methods face significant challenges in cross-domain applications, as further evidenced by our analysis in Table~\ref{tab:in_out_domain}.





\begin{table}[t]
    \centering
    \resizebox{0.7\columnwidth}{!}{%
    \begin{tabular}{l|cccc}
    \multicolumn{5}{c}{MT (Kashmiri to English)} \\
    \toprule
    \#Iter & BLEU & BERT & ChrF++ & NoAct\% \\ 
    \midrule
      Iter 0 &5.92 & 87.24 & 26.23 &  17.70 \\
     Iter 1  & 11.01 & 89.18 & 33.05& 79.71   \\
      Iter 2  & 11.87 & 89.36 & 33.41& 90.67   \\
      Iter 3  &  12.09 & 89.41 & 33.41& 95.28 \\
      Iter 4   & 12.26 & 89.45 & 33.43 & 97.21  \\
      Iter 5   & 12.36& 89.47 & 33.39&  -\\

      \bottomrule

    \end{tabular}   
    }
\vspace*{-2mm}
    \caption{The influence of 5 iterations on main results of MT using Prog-Refine (Zero-shot Act.) on CONV test set. NoAct\%: The percentage of utterances requiring no refinement, as indicated by `NoAction'. }
    \vspace*{-2mm}
    \label{tab:mt_result_iter}
\end{table}

\subsection{Ablation Study}

\begin{table}[t]
    \centering
         \resizebox{0.7\columnwidth}{!}{%
    \begin{tabular}{l|ccc}
    \multicolumn{4}{c}{MT (Kashmiri to English)} \\
    \toprule
     & BLEU & BERT & ChrF++ \\ 
           \midrule
           Initial &5.92& 87.24 & 26.23 \\
    \midrule
      Edit: DEL, INS  &12.36& 89.47 & 33.39 \\
      Edit: DEL   & 12.27 & 89.42 & 33.21\\
      Edit: INS   & 12.18 & 89.45 & 33.42 \\
      \midrule
      Unordered: DEL, INS  & 7.12  & 87.86 & 29.21 \\
      Unordered: DEL   & 6.52 & 87.51 & 26.46 \\
      Unordered: INS   & 7.14 & 88.04 & 30.38 \\

      \bottomrule
    \end{tabular}    
    }
\vspace*{-3mm}
    \caption{The results of MT using Prog-Refine (Zero-shot Act.) on CONV test set at 5th iteration with different types of actions. Edit: Actions are generated based on edit distance. Unordered: Actions without any specific order. INS: Insertion. DEL: Deletion.}
    \vspace*{-3mm}
    \label{tab:action_type}
\end{table}

\paragraph{Refinement Iterations.} In Table~\ref{tab:mt_result_iter}, we observe that Prog-Refine significantly improves the initial output generated by the Generator. However, it only demonstrates marginal improvements in the subsequent outputs from the Interpreter, even after four additional iterations. We hypothesize that this limited improvement may be attributed to training the model solely on synthetic data generated by the Generator, so the action distribution might be different to the ones for modifying the output of the Interpreter in the subsequent iterations. 

\paragraph{Action Types.} We further examine the impact of solely utilizing one type of action and the influences of disregarding the sequence of these actions. In the setting with unordered actions, oracle actions are generated by simply contrasting the differences within two sentences' unordered sets of words. As depicted in Table~\ref{tab:action_type}, the \textit{Delete} and \textit{Insert} actions, when used individually, can deliver performance metrics on par with when they are combined. However, ignoring the order of actions can lead to a substantial decline in the refinement performance. This highlights that LLM editing methods like PiVe, which utilize unordered insertions, are not optimally suited for our tasks. Further analysis is in Appendix \ref{app:action}.

\paragraph{Domain Discrepancy.} As shown in Table~\ref{tab:in_out_domain}, a domain shift dramatically impacts the performance of flan-T5 and Self-Correct. While both baseline models show markedly superior performance on the in-domain test set relative to our model, ours either surpasses or equals their performance in the cross-domain MT and AMR-to-Text test sets. This disparity in performance is likely due to the smaller models' limited cross-domain generalization. Similarly, in MT tasks, our preliminary experiments show that fine-tuned mT5 achieves 30 points of BLEU on the in-domain test but only 16 and 13 on out-of-domain tests. For further details on domain discrepancies, see Appendix \ref{app:domain}.
\begin{table}[t]
\small
    \centering
         \resizebox{0.75\columnwidth}{!}{%
    \begin{tabular}{l|ccc}
    \multicolumn{4}{c}{ LF-to-Text (AMR to English)} \\
    \toprule
       Method  & BLEU& BERT & ChrF++   \\
    \midrule
      Fine-tuned flan-T5 &  \textbf{34.63}  & \textbf{95.05} & \textbf{66.97}  \\
      GPT-3.5                       &  &&  \\
      \ \ \ \ Initial               &  19.67 & 92.10 &  55.98 \\
      \ \ \ \ Self-Correct           & 34.49 & 94.68 & 66.81\\
      \ \ \ \ Self-Refine           &  16.16 & 91.08 &  52.78\\
      \ \ \ \ Prog-Refine    & 29.12 & 94.01 & 64.85 \\
      \bottomrule    
    \end{tabular}    
    }
\vspace*{-3mm}
    \caption{LF-to-Text results using Prog-Refine (Zero-shot Act.) on the in-domain LDC test.}
    \vspace*{-3mm}
    \label{tab:in_out_domain}
\end{table}

\paragraph{Adversarial In-context Learning.} Table~\ref{tab:adv_ic} indicates 0.0 for no corruption and 1.0 for complete discarding of exemplar actions, leaving only ${(\vx_{j},\vy^{*}_{j},\vy_{j})}_{j=1}^{n}$. Rates between 0.0 and 1.0 represent partial corruption of Oracle actions. The results suggest that neither full application nor total corruption of Oracle actions is optimal. However, partial corruption leads to improved performance. Additionally, across all corruption rates, few-shot settings consistently outperform zero-shot settings.

\begin{table}[t]
\small
    \centering
         \resizebox{0.6\columnwidth}{!}{%
    \begin{tabular}{l|ccc}
    \multicolumn{4}{c}{LF-to-Text (AMR to English)} \\
    \toprule
       Rate  & BLEU& BERT & ChrF++   \\
    \midrule
      0.0       & 12.06  & 89.31 & 46.23 \\
      0.2     &  12.35 & \textbf{89.36} &  46.49 \\
      0.5           & \textbf{13.64} & 89.27 & \textbf{47.69}\\
      1.0           &  11.97 & 89.32 &  46.13 \\
      \bottomrule    
    \end{tabular}    
    }
\vspace*{-3mm}
    \caption{LF-to-Text results using Prog-Refine (Few-shot Act.) vary with different corruption probabilities for the action sequence in the adversarial in-context examples used for the Interpreter.}
    \vspace*{-3mm}
    \label{tab:adv_ic}
\end{table}


\section{Conclusions}
We present a programmer-interpreter method that iteratively refines LLM outputs using edit actions from a fine-tuned programmer and an LLM interpreter. Our approach combines the task-specific encoding capacity of a fine-tuned model with the domain generalization strength of the LLM, incorporating specifically designed actions for text generation. The experiments confirm its efficacy, showing significant improvements in LLM-generated text quality for low-resource MT and LF-to-Text tasks. Moreover, our approach outperforms established baselines in cross-domain scenarios. 


\section{Limitations}
This work has two primary limitations. First, in in-domain tests, our approach does not outperform smaller models, such as mT5 and flan-T5. Considering the performance improvements we observed when using ORACLE actions, we believe there is substantial potential to further enhance our method for text generation in the in-domain evaluation setting. Second, our approach requires internet transmission of prompt instructions to the servers of ChatGPT. This could potentially lead to a risk of privacy leakage, which is a critical concern in data-sensitive applications.


\bibliography{custom}

\begin{thebibliography}{18}
\expandafter\ifx\csname natexlab\endcsname\relax\def\natexlab#1{#1}\fi

\bibitem[{Chung et~al.(2022)Chung, Hou, Longpre, Zoph, Tay, Fedus, Li, Wang, Dehghani, Brahma et~al.}]{chung2022scaling}
Hyung~Won Chung, Le~Hou, Shayne Longpre, Barret Zoph, Yi~Tay, William Fedus, Eric Li, Xuezhi Wang, Mostafa Dehghani, Siddhartha Brahma, et~al. 2022.
\newblock Scaling instruction-finetuned language models.
\newblock \emph{arXiv preprint arXiv:2210.11416}.

\bibitem[{Gala et~al.(2023)Gala, Chitale, AK, Doddapaneni, Gumma, Kumar, Nawale, Sujatha, Puduppully, Raghavan et~al.}]{gala2023indictrans2}
Jay Gala, Pranjal~A Chitale, Raghavan AK, Sumanth Doddapaneni, Varun Gumma, Aswanth Kumar, Janki Nawale, Anupama Sujatha, Ratish Puduppully, Vivek Raghavan, et~al. 2023.
\newblock Indictrans2: Towards high-quality and accessible machine translation models for all 22 scheduled indian languages.
\newblock \emph{arXiv preprint arXiv:2305.16307}.

\bibitem[{Han et~al.(2023)Han, Collier, Buntine, and Shareghi}]{han2023pive}
Jiuzhou Han, Nigel Collier, Wray Buntine, and Ehsan Shareghi. 2023.
\newblock Pive: Prompting with iterative verification improving graph-based generative capability of llms.
\newblock \emph{arXiv preprint arXiv:2305.12392}.

\bibitem[{Haroutunian et~al.(2023)Haroutunian, Li, Galescu, Cohen, Tumuluri, and Haffari}]{haroutunian2023reranking}
Levon Haroutunian, Zhuang Li, Lucian Galescu, Philip Cohen, Raj Tumuluri, and Gholamreza Haffari. 2023.
\newblock Reranking for natural language generation from logical forms: A study based on large language models.
\newblock \emph{arXiv preprint arXiv:2309.12294}.

\bibitem[{Lai et~al.(2023)Lai, Ngo, Veyseh, Man, Dernoncourt, Bui, and Nguyen}]{lai2023chatgpt}
Viet~Dac Lai, Nghia~Trung Ngo, Amir Pouran~Ben Veyseh, Hieu Man, Franck Dernoncourt, Trung Bui, and Thien~Huu Nguyen. 2023.
\newblock Chatgpt beyond english: Towards a comprehensive evaluation of large language models in multilingual learning.
\newblock \emph{arXiv preprint arXiv:2304.05613}.

\bibitem[{Madaan et~al.(2023)Madaan, Tandon, Gupta, Hallinan, Gao, Wiegreffe, Alon, Dziri, Prabhumoye, Yang et~al.}]{madaan2023self}
Aman Madaan, Niket Tandon, Prakhar Gupta, Skyler Hallinan, Luyu Gao, Sarah Wiegreffe, Uri Alon, Nouha Dziri, Shrimai Prabhumoye, Yiming Yang, et~al. 2023.
\newblock Self-refine: Iterative refinement with self-feedback.
\newblock \emph{arXiv preprint arXiv:2303.17651}.

\bibitem[{Papineni et~al.(2002)Papineni, Roukos, Ward, and Zhu}]{papineni2002bleu}
Kishore Papineni, Salim Roukos, Todd Ward, and Wei-Jing Zhu. 2002.
\newblock Bleu: a method for automatic evaluation of machine translation.
\newblock In \emph{Proceedings of the 40th annual meeting of the Association for Computational Linguistics}, pages 311--318.

\bibitem[{Pillutla et~al.(2021)Pillutla, Swayamdipta, Zellers, Thickstun, Welleck, Choi, and Harchaoui}]{DBLP:conf/nips/PillutlaSZTWCH21}
Krishna Pillutla, Swabha Swayamdipta, Rowan Zellers, John Thickstun, Sean Welleck, Yejin Choi, and Za{\"{\i}}d Harchaoui. 2021.
\newblock {MAUVE:} measuring the gap between neural text and human text using divergence frontiers.
\newblock In \emph{Advances in Neural Information Processing Systems (NeurIPS)}, pages 4816--4828.

\bibitem[{Popovi{\'c}(2017)}]{popovic2017chrf++}
Maja Popovi{\'c}. 2017.
\newblock chrf++: words helping character n-grams.
\newblock In \emph{Proceedings of the second conference on machine translation}, pages 612--618.

\bibitem[{Reed and de~Freitas(2016)}]{DBLP:journals/corr/ReedF15}
Scott~E. Reed and Nando de~Freitas. 2016.
\newblock Neural programmer-interpreters.
\newblock In \emph{International Conference on Learning Representations (ICLR)}.

\bibitem[{Sutskever et~al.(2014)Sutskever, Vinyals, and Le}]{sutskever2014sequence}
Ilya Sutskever, Oriol Vinyals, and Quoc~V Le. 2014.
\newblock Sequence to sequence learning with neural networks.
\newblock \emph{Advances in neural information processing systems}, 27.

\bibitem[{Vu and Haffari(2018)}]{vu2018automatic}
Thuy Vu and Gholamreza Haffari. 2018.
\newblock Automatic post-editing of machine translation: A neural programmer-interpreter approach.
\newblock In \emph{Proceedings of the 2018 conference on empirical methods in natural language processing}, pages 3048--3053.

\bibitem[{Wang et~al.(2023)Wang, Xixu, Hou, Chen, Zheng, Wang, Yang, Ye, Huang, Geng et~al.}]{wang2023robustness}
Jindong Wang, HU~Xixu, Wenxin Hou, Hao Chen, Runkai Zheng, Yidong Wang, Linyi Yang, Wei Ye, Haojun Huang, Xiubo Geng, et~al. 2023.
\newblock On the robustness of chatgpt: An adversarial and out-of-distribution perspective.
\newblock In \emph{ICLR 2023 Workshop on Trustworthy and Reliable Large-Scale Machine Learning Models}.

\bibitem[{Welleck et~al.(2022)Welleck, Lu, West, Brahman, Shen, Khashabi, and Choi}]{welleck2022generating}
Sean Welleck, Ximing Lu, Peter West, Faeze Brahman, Tianxiao Shen, Daniel Khashabi, and Yejin Choi. 2022.
\newblock Generating sequences by learning to self-correct.
\newblock \emph{arXiv preprint arXiv:2211.00053}.

\bibitem[{Xue et~al.(2021)Xue, Constant, Roberts, Kale, Al-Rfou, Siddhant, Barua, and Raffel}]{xue2021mt5}
Linting Xue, Noah Constant, Adam Roberts, Mihir Kale, Rami Al-Rfou, Aditya Siddhant, Aditya Barua, and Colin Raffel. 2021.
\newblock mt5: A massively multilingual pre-trained text-to-text transformer.
\newblock In \emph{Proceedings of the 2021 Conference of the North American Chapter of the Association for Computational Linguistics: Human Language Technologies}, pages 483--498.

\bibitem[{Yang et~al.(2023)Yang, Jin, Tang, Han, Feng, Jiang, Yin, and Hu}]{yang2023harnessing}
Jingfeng Yang, Hongye Jin, Ruixiang Tang, Xiaotian Han, Qizhang Feng, Haoming Jiang, Bing Yin, and Xia Hu. 2023.
\newblock Harnessing the power of llms in practice: A survey on chatgpt and beyond.
\newblock \emph{arXiv preprint arXiv:2304.13712}.

\bibitem[{Zhang et~al.()Zhang, Kishore, Wu, Weinberger, and Artzi}]{zhangbertscore}
Tianyi Zhang, Varsha Kishore, Felix Wu, Kilian~Q Weinberger, and Yoav Artzi.
\newblock Bertscore: Evaluating text generation with bert.
\newblock In \emph{International Conference on Learning Representations}.

\bibitem[{Zhuo et~al.(2023)Zhuo, Li, Huang, Shiri, Wang, Haffari, and Li}]{zhuo2023robustness}
Terry~Yue Zhuo, Zhuang Li, Yujin Huang, Fatemeh Shiri, Weiqing Wang, Gholamreza Haffari, and Yuan-Fang Li. 2023.
\newblock On robustness of prompt-based semantic parsing with large pre-trained language model: An empirical study on codex.
\newblock In \emph{Proceedings of the 17th Conference of the European Chapter of the Association for Computational Linguistics}, pages 1090--1102.

\end{thebibliography}

\clearpage

\appendix

\section{Appendix}
\label{sec:appendix}




\subsection{Prompt Example for Editing Text}

\begin{figure}[ht]
    \centering
    \includegraphics[width=\columnwidth]{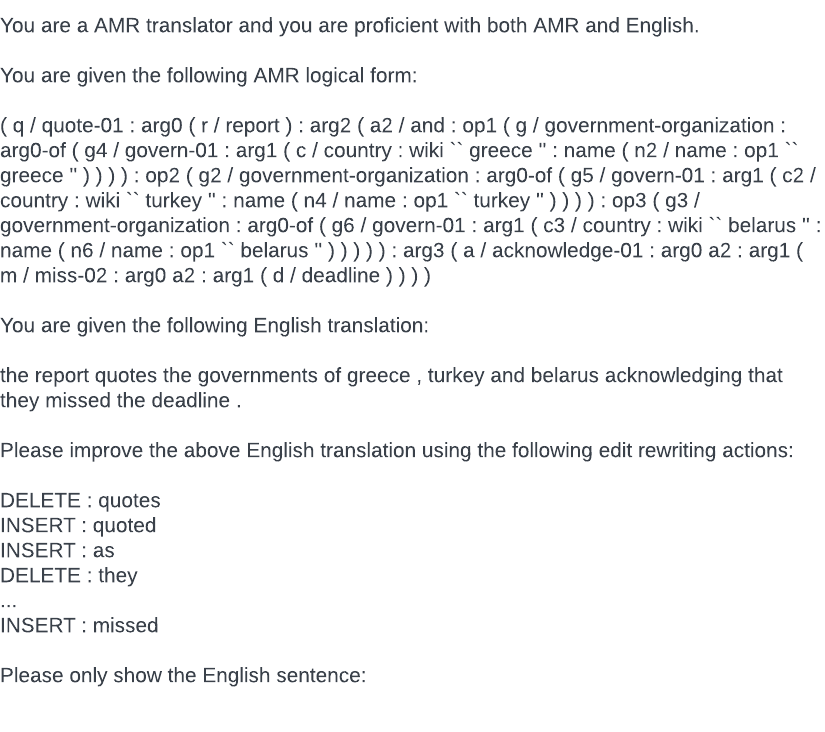}
    \caption{The zero-shot exemplary prompt for LF-to-Text.}
    \label{fig:example_zeroshot}
\end{figure}

\begin{figure}[ht]
    \centering
    \includegraphics[width=\columnwidth]{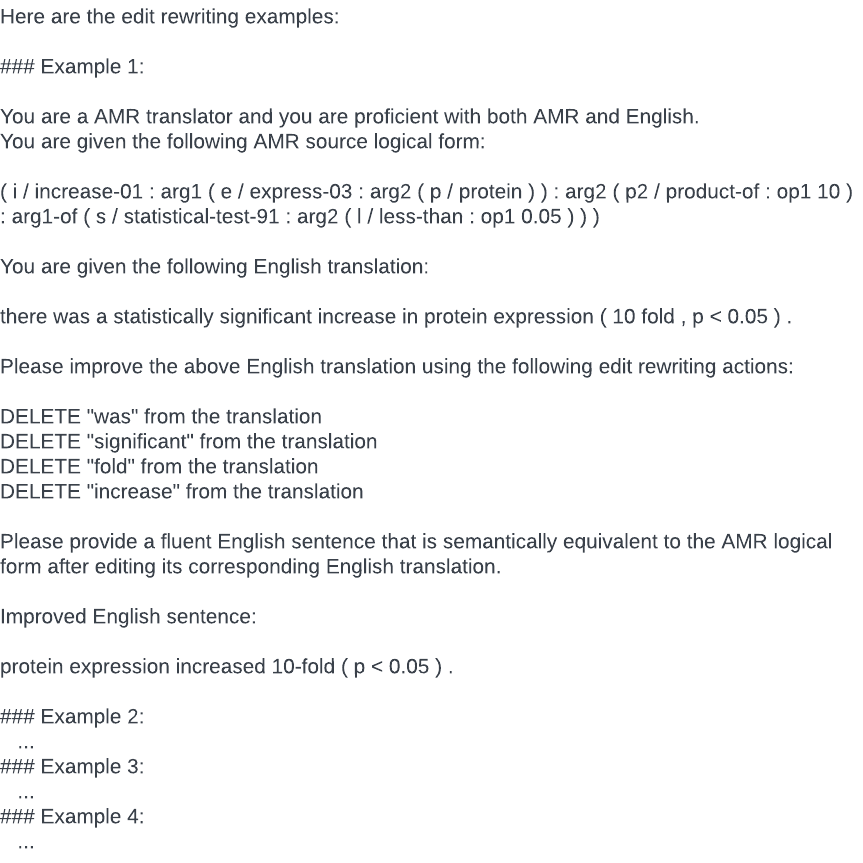}
    \caption{The few-shot exemplary prompt for LF-to-Text.}
    \label{fig:examplar_adv}
\end{figure}

Figures~\ref{fig:example_zeroshot} and~\ref{fig:examplar_adv} depict the exemplary zero/few-shot prompt employed in LF-to-Text.

\subsection{Adaption of Self-Corrector}
In our experiment, we adapted the implementation of the Self-Corrector to better suit our specific requirements. To customize it for our context, we constructed the training set for the Self-Corrector's Interpreter as follows: the input consists of a concatenation of Kashrimi/AMR, text produced by the Generator, and edit actions. The output, on the other hand, is the ground truth text. For a fair comparison with our approach and to minimize training and data collection expenses, models are trained only during the first iteration. Additionally, the generation of the training set solely utilizes text from the Generator in the initial iteration, without using text from the Interpreter in subsequent refinement iterations.

\subsection{Measures of Domain Discrepancy}
\label{app:domain}
Tables \ref{tab:mt-domain-difference} and \ref{tab:amr-domain-difference} present domain discrepancies for the training/development/testing sets for the MT and LF-to-text generation tasks. The domain discrepancy measures include the KL-divergence (based on the unigram distributions) and MAUVE \cite{DBLP:conf/nips/PillutlaSZTWCH21}. KL-divergence scores are higher when two distributions are more different from each other. MAUVE scores, which have a range (0,1), are lower when two distributions are more different from each other.

\begin{table}[t]
    \centering
    \begin{tabular}{|l|c|c|}
       \hline
       \textbf{splits compared}  & \textbf{KL-div $\downarrow$} &  \textbf{MAUVE $\uparrow$} \\
       \hline
        train, dev & $2.23$ & $0.006$ \\
        dev, test$_{gen}$ & $1.97$ & $0.231$ \\
        train, test$_{gen}$ & $1.94$ & $0.005$ \\ 
        dev, test$_{conv}$ & $2.97$ & $0.040$ \\
        train, test$_{conv}$ & $2.98$ & $0.007$ \\ 
    \hline
    \end{tabular}
    \caption{Measures of domain difference across different splits of the machine translation datasets. KL-divergence scores are calculated for the English sentences in each data split, with additive smoothing ($\alpha = 1 \times 10^{-4}$). For MAUVE, 5000 sentences are sampled from the training set.}
    \label{tab:mt-domain-difference}
\end{table}

\begin{table}[t]
    \centering
    \begin{tabular}{|l|c|c|}
       \hline
       \textbf{splits compared}  & \textbf{KL-div $\downarrow$} &  \textbf{MAUVE $\uparrow$} \\
       \hline
        train, dev & $2.00$ & $0.512$ \\
        dev, test$_{i.d.}$ & $2.39$ & $0.327$ \\
        train, test$_{i.d.}$ & $1.97$ & $0.342$ \\
        dev, test$_{bio}$ & $6.01$ & $0.004$ \\
        train, test$_{bio}$ & $5.48$ & $0.004$ \\ 
    \hline
    \end{tabular}
    \caption{Measures of domain difference across different splits of the AMR dataset.   KL-divergence scores are calculated for the English sentences in each data split, with additive smoothing ($\alpha = 1 \times 10^{-4}$). For MAUVE, 5000 sentences are sampled from the training set.}
    \label{tab:amr-domain-difference}
\end{table}

Based on Table \ref{tab:mt-domain-difference}, we observe that the domain of test-gen is closer to the training set compared to that of the test-conv. This is pronounced in higher KL-divergence and lower MAUVE numbers for the test-conv compared to test-gen, with respect to the training set.  

Based on Table \ref{tab:amr-domain-difference} , we observe a higher difference for the domain of the biology-AMR test compared to the LDC2.0-AMR test set, with respect to  the training/development sets of the LDC2.0-AMR dataset. This is pronounced in larger KL divergence and lower MAUVE numbers compared to those for the LDC2.0-AMR test set. 

\subsection{F1 Definition for Action Prediction}
\begin{align}
F1 &= 2 \times \frac{P_{act}\times R_{act}}{P_{act}+R_{act}}
\end{align}

\noindent Here, $P_{act}$ represents action precision, defined as the ratio of predicted actions present in the reference action sequence to the total number of predicted actions. $R_{act}$ denotes action recall, which is the ratio of predicted actions that appear in the reference action sequence to the total number of actions in the reference sequence. The F1 score, thus, provides a harmonious mean of these two metrics.

\subsection{F1 for Action Prediction}
\label{app:action}
Table~\ref{tab:my_label} reveals that predicting INSERT actions is a relatively easier task compared to predicting DELETE actions. This observation is reasonable since the Programmer only needs to learn how to DELETE words from the text with a fixed vocabulary, whereas, for INSERT actions, the Programmer must learn to INSERT arbitrary words.
\begin{table}[t]
    \centering
     \resizebox{0.65\columnwidth}{!}{%
    \begin{tabular}{l|ccc}
    \toprule     
           &  INSERT & DELETE & Total \\
      \midrule     
       MT  &  33.64  & 83.73  &    62.57     \\  
       NLG  &   24.52 & 60.48  & 44.90       \\  
    \toprule 

    \end{tabular}
    }
    \vspace*{-2mm}
    \caption{The F1 scores of comparing the predicted actions with the ORACLE actions in the GEN test set.}
    \label{tab:my_label}
        \vspace{-3mm}
\end{table}

\subsection{Comparing GPT-4 and GPT-3.5 as Interpreters}
\begin{table}[t]
\small
\centering
\begin{tabular}{l|ccc}
\multicolumn{4}{c}{LF-to-Text (AMR to English)} \\
\toprule
& BLEU & BERT & ChrF++ \\
\midrule
GPT-3.5-turbo-16k & 11.43 & 89.30 & 45.44 \\
GPT-4-turbo & 11.72 & 89.36 & 45.58 \\
\bottomrule
\end{tabular}
\caption{LF-to-Text results of Prog-Refine (Zero-shot Act.) in zero-shot setting with different LLMs as Interpreters.}
\label{tab:gpt_compare}
\end{table}

Table~\ref{tab:gpt_compare} illustrates the performance differences in the LF-to-Text task when using GPT-4 and GPT-3.5 as Interpreters for Prog-Refine (Zero-shot Act.). While GPT-4 offers a slight performance boost, the improvement is not substantial, amounting to only a 0.3 increase in BLEU score. Moreover, this comes at a higher cost of 0.06 per 1000 characters, compared to 0.0015 for GPT-3.5.

\end{document}